%% file: ijcai22.tex
\documentclass{article}
\usepackage{ijcai22}
\usepackage{times}
\usepackage{soul}
\usepackage{url}
\usepackage[hidelinks]{hyperref}
\usepackage[utf8]{inputenc}
\usepackage[small]{caption}
\usepackage{graphicx}
\usepackage{amsmath}
\usepackage{amsthm}
\usepackage{natbib}
\usepackage{booktabs}
\usepackage{algorithm}
\usepackage{algpseudocode}
\usepackage{bm}
\usepackage{times}
\usepackage{latexsym}
\usepackage{amsmath,amsfonts,amssymb,bbm,epsfig,bm}
\usepackage{balance}
\usepackage{booktabs}
\usepackage{multirow}
\usepackage{array}
\usepackage{url}
\usepackage{tikz}
\usepackage{makecell}
\usepackage{subfigure}
\usepackage{framed}
\usepackage{pstricks}
\usepackage{graphicx}
\usepackage{color}
\usepackage{enumerate}
\usepackage{arydshln}
\usepackage{mathrsfs}
\usepackage{pifont}
\usepackage{xspace}

\usepackage{nicefrac}
\usepackage{amsthm}
\usepackage{mathtools}
\usetikzlibrary{arrows}
\usepackage{enumitem}
\usepackage{hyperref}
\usepackage{caption}

\newcommand{\vect}[1]{\mathbf{#1}}

\renewcommand{\algorithmiccomment}[1]{\bgroup\hfill $\triangleright$ ~#1\egroup}

\newcommand{\ours}{\textsc{Match-Tuning}\xspace}

\renewcommand{\algorithmiccomment}[1]{\bgroup\hfill $\triangleright$ ~#1\egroup}

\title{Robust Fine-tuning via Perturbation and Interpolation \\
from In-batch Instances}
\author{
Shoujie Tong\textsuperscript{\rm1 $*$}, Qingxiu Dong\textsuperscript{\rm1 $*$}, Damai Dai\textsuperscript{\rm 1}, Yifan Song\textsuperscript{\rm 1}, \\Tianyu Liu\textsuperscript{\rm 2}, Baobao Chang\textsuperscript{\rm 1}, Zhifang Sui\textsuperscript{\rm 1}\\ 
\textsuperscript{\rm 1} Key Laboratory of Computational Linguistics, Peking University 
\textsuperscript{\rm 2} Tencent Cloud Xiaowei \\
  \texttt{ \{tong,dqx\}@stu.pku.edu.cn}
  \texttt{ \{daidamai,yfsong,chbb,szf\}@pku.edu.cn}
  \texttt{ \{rogertyliu\}@tencent.com}
}
\begin{document}

\maketitle
\renewcommand{\thefootnote}{\fnsymbol{footnote}}
\footnotetext[1]{Equal contribution.}
\renewcommand{\thefootnote}{\arabic{footnote}}
\begin{abstract}
Fine-tuning pretrained language models (PLMs) on downstream tasks has become common practice in natural language processing. However, most of the PLMs are vulnerable, e.g., they are brittle under adversarial attacks or imbalanced data, which hinders the application of the PLMs on some downstream tasks, especially in safe-critical scenarios. In this paper, we propose a simple yet effective fine-tuning method called \ours{} to force the PLMs to be more robust. For each instance in a batch, we involve other instances in the same batch to interact with it. To be specific, regarding the instances with other labels as a perturbation, \ours{} makes the model more robust to noise at the beginning of training. While nearing the end, \ours{} focuses more on performing an interpolation among the instances with the same label for better generalization. Extensive experiments on various tasks in GLUE benchmark show that \ours{} consistently outperforms the vanilla fine-tuning by $1.64$ scores. Moreover, \ours{} exhibits remarkable robustness to adversarial attacks and data imbalance.\footnote{Our code is available at \url{https://github.com/tongshoujie/MATCH-TUNING}}
\end{abstract}

\section{Introduction}

Pretrained language models (PLMs) have contributed to striking success in natural language processing (NLP). Simultaneously, fine-tuning has been a common practice to employ PLMs for downstream natural language understanding tasks.
However, recent work shows that vanilla fine-tuning methods may lead to vulnerable models~\citep{rxf}. This long-standing problem hinders the model performance and makes fine-tuned PLMs vulnerable to adversarial attacks and spurious bias~\citep{branco2021shortcutted,2019dont}. As a result, it limits the application of the PLMs on some downstream tasks, especially in some real-world scenarios where robustness is especially required.

\input{float/figure-process}

To alleviate this problem, various fine-tuning approaches have been proposed. For instance, SMART~\citep{smart} and R3F ~\citep{rxf} introduce regularizations to the noise applied to the original pretrained representations. ChildTuning ~\citep{xu2021raise} updates the child network during fine-tuning via strategically masking out the gradients of the non-child network. However, most of them focus on improving the generalizing robustness or the adaptive robustness, while the robustness to adversarial attacks and spurious correlations remains challenging.

Inspired by contrastive learning with in-batch instances~\citep{gao2021simcse,fan2021does}, we propose \ours{} that utilizes in-batch instances dynamically for robust fine-tuning. 
In \ours{}, we convert each instance representation in a batch by fusing the representations of other instances in the same batch. 
More concretely, \ours{} first calculates the similarities between the PLM outputs of instances in a batch to form a matching matrix.
Then, we fuse the PLM representations according to the matching matrix to form new representations for this batch. 
Finally, we use the new representations for prediction.

\ours{} works by adaptively determining how to utilize the in-batch instances during the whole training procedure. 
As shown in Fig.~\ref{fig:process}, at the beginning of training, regarding the instances with other labels as a perturbation, \ours{} urges the PLM to converge to a more flat local minimum for better robustness. 
While nearing the end, \ours{} focuses more on performing an interpolation among the instances with the same label for better generalization. 
In this manner, \ours{} reduces the vulnerability of models and improves their general performance.

We conduct a comprehensive evaluation of \ours{} on the GLUE benchmark. 
The results show that our method outperforms the vanilla fine-tuning by $1.64$ scores on average. 
In addition, our method outperforms vanilla fine-tuning by $4.11$ average scores on advGLUE, and yields a great improvement for label noise or data imbalance, which shows our overwhelming robustness over vanilla fine-tuning. 

Our main contributions are summarized as follows:
\begin{itemize}
    \item We propose an adaptive fine-tuning method called \ours{} to train robust models, where instances in the same batch will interact with each other. 
    \item \ours{} reduces the vulnerability of models and outperforms the vanilla fine-tuning by $1.64$ scores on the GLUE benchmark.
    \item Our method manifests extraordinary robustness to various scenarios, including adversarial attacks, spurious biases, and data imbalance. 
\end{itemize}

\section{Related Work}\label{related}
The vanilla fine-tuning simply adapts PLMs to the task-specific inputs and outputs, and fine-tunes all the parameters in an end-to-end manner~\citep{bert,roberta}. The token representations or the representation of a special token~(e.g., [CLS]) is directly fed into an output layer for tagging or classification, respectively. 
This manner has been shown to produce biased models that are vulnerable to adversarial attacks and noisy data~\citep{rxf,2019dont}. 

In the past years, numerical variants like ChildTuning and R3F are proposed to conduct more trustworthy and effective fine-tuning~\citep{mixout, xu2021raise, rxf}. 
FreeLB~\citep{zhu2019freelb} adds perturbations to continuous word embedding by using a gradient method and minimizes the resultant adversarial risk. 
Moreover, ~\citet{smart} introduce a regularization to encourage the model output not to change much when injecting a small perturbation to the input. 
\citet{rxf} simply regularize the model against the parametric noise. 
The literature provides strong insights that proper perturbation on the PLM outputs has great potential in enforcing the model smoothness and robustness. 

Recently, in-batch learning successes in many fields. 
\citet{liu2019knowledge,yao2021instance} suggest that the unsupervised instances are helpful to learning the classifier in computer vision tasks. 
For NLP, contrastive learning with in-batch instances also improves the task-specific representation~\citep{gao2021simcse} and adversarial robustness~\citep{fan2021does}.
Inspired by this, we propose \ours{} to utilize in-batch instances dynamically for robust fine-tuning. 
Different from previous work, \ours{} no longer needs label information or the specification of negative and positive instances in advance. 
It performs automatic instance interaction, which applies to most existing pretrained models.

\section{Method}
\ours{} derives a composite representation for each instance in a batch by fusing the representations of other instances in the same batch, which are deemed as adaptive noise. 
From experiments, we find that the adaptive noise functions as an in-batch perturbation in the initial stage of training, and then gradually transits to an in-batch interpolation among ``positive'' instances that share the same label. 
In addition, we show that \ours{} helps the model to escape the sharp local minima through qualitative analysis. 

\subsection{Overview of \ours{}}
In the batched gradient descent, the data points in a batch are formulated as $\{(x_{i}, y_{i}) | i=1,...,n\}$, where $n$ denotes the batch size. 
Throughout our paper, $x_i$ represents a textual input, e.g., a single sentence or a sentence pair, while $y_i$ denotes a discrete label or a continuous number for classification and regression tasks, respectively.
We use $h$ and $\boldsymbol{\theta}$ to represent a PLM that extracts contextualized features from $x_i$ and its parameters. 
Similarly, the task-specific classifier and its parameters are denoted by $f$ and $\boldsymbol{\psi}$. 
Letting $L$ denote the task-specific loss function, we compute the mini-batch gradient $\boldsymbol{g}$ in the vanilla fine-tuning as follows:

\begin{align}
\boldsymbol{g}=\frac{1}{n} \nabla_{\boldsymbol{\theta}, \boldsymbol{\psi}} \sum_{i=1}^{n} L\left(f(h(x^{(i)};\boldsymbol{\theta});\boldsymbol{\psi}), y^{(i)}\right).
\end{align}

To apply adaptive weights to the instances in a batch, we introduce the matching matrix $M$, where each element indicates the pair-level similarity between in-batch instance representations given by a PLM. 
The matrix $M$ is given by
\begin{align}
% M = \sum_{j=1}^{n}\frac{exp(x^{i} x^{j})}{\sum_{k=1}^{n} exp(x^{i} x^{k})}{x^{j}}
M_{i,j} = \frac{\operatorname{exp}\left(h(x^{(i)};\boldsymbol{\theta}) h(x^{(j)};\boldsymbol{\theta})\right)}{\sum_{k=1}^{n} \operatorname{exp}\left(h(x^{(i)};\boldsymbol{\theta}) h(x^{(k)};\boldsymbol{\theta})\right)}.
\end{align}

Note that the overhead to compute ${M}_{i,j}$ is small since we can directly reuse $h(x;\boldsymbol{\theta})$, the outputs of the PLM. 
Then, to produce a robust representation for an instance, we derive a composite representation $\mathbf{z}^{(i)}$ from $h(x^{(i)};\boldsymbol{\theta})$:
\begin{align}\label{algo:transed}
 \mathbf{z}^{(i)}=\sum_{j=1}^{n}M_{i,j}h(x^{(j)};\boldsymbol{\theta}).
\end{align}
Then, $\mathbf{z}^{(i)}$ serves as a drop-in replacement is simple and easy-to-use.
for $h(x^{(i)};\boldsymbol{\theta})$ in the vanilla fine-tuning and the mini-batch gradient $\boldsymbol{g}^{\prime}$ in \ours{} is computed as follows:
\begin{align}
\boldsymbol{g}^{\prime}=\frac{1}{n} \nabla_{\boldsymbol{\theta},\boldsymbol{\psi}} \sum_{i=1}^{n} L\left(f(\mathbf{z}^{(i)};\boldsymbol{\psi}), y^{(i)}\right).
\end{align}

\input{float/table-main}

\subsection{Qualitative Understanding of \ours{}} \label{sec:theory}
\input{float/figure-minima}

According to our observations in the experiments, \ours{} converges faster and reaches a better global minima compared with vanilla fine-tuning\footnote{The comparison of training loss is provided in Appendix D}.
We provide a qualitative viewpoint to understand how \ours{} works stemming from the notions of perturbation and interpolation.

We first introduce the shape of local minima. 
The loss surface of deep neural networks tends to have various local minima as illustrated in Fig.~\ref{fig:minina}. 
Sharp local minima are where the loss in a small neighborhood increase rapidly while flat local minima are where the loss varies slowly in a relatively large neighborhood. 
Sufficient literature has proved that flat local minima usually lead to better generalization~\citep{hochreiter1995simplifying,dinh2017sharp}. 
In addition, under a Bayesian perspective, the noise in gradient could drive model away from sharp minima~\citep{smith2018don}. 

\ours{} follows this gradient noise addition routine, but in a different way. To better understand the mechanism of \ours{}, we visualize the the values of the ``positive'' instances (with the same label) and ``negative'' instances (with other labels) in the matching matrix. 
Fig.~\ref{fig:adaptive} depicts the change of matching matrix in \ours{}. We sum up the values of all instances with the same label and show the cumulative values for ``positive'' instances on Fig.~\ref{fig:adaptive}. 
The ``negative'' instances share the same setting.

%provides an adaptive process by the matching matrix. 
\input{float/figure-adaptive} 
\paragraph{In-batch Perturbation} 
In the initial stage, \ours{} works by performing in-batch perturbation on the original outputs of PLM. 
As shown in Fig.~\ref{fig:adaptive}, instances from the same class and that from other classes are similar in the matching matrix. 
Therefore, any other instance in the same batch is close to a tiny perturbation on the output of PLM. 
If the PLM provides vulnerable representations, the perturbed representations will break down easily. 
Therefore, the early stage in \ours{} encourages the PLM to generate a more robust representation for each instance and converge to a more flat local minimum. 

\paragraph{In-batch Interpolation}
During the whole training process, the model will gradually learn to distinguish the representations of the positive (same label) and negative (other labels) instances in a batch.
Consequently, as training proceeds, the portion of negative instances is getting smaller in the match matrix and can hardly influence the composite representation in the late stage of training. 
In this moment, \ours{} tends to interpolate the representations of the positive instances. 
We also observe more representative positive instances will contribute more to the final composite representation. 
As illustrated in Fig.~\ref{fig:process}, the late stage in \ours{} encourages composite representations to be grouped into clusters according to their real labels. 

\section{Experiments}
We conducted extensive experiments on various downstream tasks to evaluate the general performance and robustness of \ours{}.
For simplicity, in the rest of the paper, we denote other instances in the same batch with the same label as the current instance by positive instances, and other instances with different labels by negative samples.

\subsection{Datasets}
Following previous work~\citep{xu2021raise,mixout}, we conduct experiments on four main datasets in \textbf{GLUE}~\citep{wang2018glue} to evaluate the general performance. 
Among them, classification task like CoLA is for linguistic acceptability, RTE is for natural language inference, and MRPC is for paraphrase identification. 
STS-B is a regression task for semantic textual similarity. 
By systematically conducting 14 kinds of adversarial attacks on representative GLUE tasks, \citet{wang2021adversarial} proposed \textbf{AdvGLUE}, a multi-task benchmark to evaluate and analyze the robustness of language models and robust training methods\footnote{Detailed information of datasets is provided in Appendix A.}.

\subsection{Experimental Setup}
We report the averaged results over $10$ random seeds. 
We conduct our experiments based on the HuggingFace transformers library\footnote{\url{https://github.com/huggingface/transformers}} and follow the default hyper-parameters and settings unless noted otherwise.
Other detailed experimental setups are presented in Appendix B.

\subsection{General Performance}
We compare \ours{} with the vanilla fine-tuning and related work on four tasks of the well-recognized benchmark, GLUE. 
And we focus on evaluating the performance of BERT-Large based models on the GLUE development set.

\paragraph{Baselines}
We compare \ours{} with the following methods.
1) \textit{Vanilla Fine-tuning}~\cite{bert}, the standard fine-tuning paradigm; 
2) \textit{Weight Decay}~\citep{daume-iii-2007-frustratingly}, which adds a regularization term to the loss function;
3) \textit{Top-$K$ Tuning}~\citep{pmlr-v97-houlsby19a}, which fine-tunes only the top-$K$ layers of the PLM;
4) \textit{Mixout}~\citep{mixout}, which stochastically replaces the parameters with their pretrained weights;
5) \textit{RecAdam}~\citep{RecAdam}, which introduces quadratic penalty and objective shifting mechanism;
6) \textit{R3F}~\citep{rxf}, which adds adversarial objectives with noise sampled from either a normal or uniform distribution;
7) \textit{ChildTuning$_F$}~\citep{xu2021raise}, which randomly masks a subset of parameters PLM during the backward process;
8) \textit{ChildTuning$_D$}~\citep{xu2021raise}, which detects the most important child network for the target task.

\paragraph{Results} We show the mean (and max) scores on GLUE Benchmark in Tab.~\ref{table:main}.
\ours significantly outperforms vanilla fine-tuning by $1.88$ average score. 
Moreover, compared with several strong tuning methods, \ours achieves the best performance on three tasks, showing its effectiveness.
Although ChildTuning$_D$ has performance on par with \ours on CoLA task (0.45 mean/-1.08 max), our method has a small computational overhead while ChildTuning$_D$ has to adopt Fisher information estimation to obtain the task-driven child network. 
Besides, \ours has consistent performance on other PLMs and we report the results in Appendix C.

Note that \ours can be integrated with other tuning methods to further boost the performance. 
We evaluate the combination of \ours and R3F, which achieves an additional improvement of 0.24 average score.

\subsection{Robustness of \ours{}}
\input{float/table-attack}
Recent work reveals that vanilla fine-tuning is deceptive and vulnerable in many aspects. 
For instance, fool the models to output arbitrarily wrong answers by perturbing input sentences in a human-imperceptible way. Real-world systems built upon these vulnerable models can be misled in ways that would have profound security concerns.
To examine the robustness of \ours{}, we design robustness evaluation tasks for three common scenarios respectively.
\subsubsection{Robustness to Adversarial Attacks}
As recent studies revealed, the robustness of fine-tuned PLMs can be challenged by carefully crafted textual adversarial examples. We systematically conduct various adversarial attack evaluations on the advGLUE benchmark. 

Tab.~\ref{table:attack} illustrates that fine-tuned models maintain vulnerabilities to adversarial attacks while our \ours{} approach alleviates this chronic problem by $4.11$ accuracy promotion on average. Compared with vanilla fine-tuning, existing methods like R3F and ChildTuning encounter $8 \sim 13$ accuracy collapse on advSST-2, while \ours{} outperforms vanilla fine-tuning by $4.11$ scores. On advMNLI, advRTE, and advQQP, \ours{} also holds a large improvement, as much as $10.79$ higher accuracy than vanilla fine-tuning. In short, compared with prior fine-tuning methods, we find that \ours{} is more robust in adapting PLMs to various tasks.

\input{float/table-noise}

\subsubsection{Robustness to Label Noise}
Nowadays, there is inevitably some noise in large-scale datasets. To explore the model robustness to noisy data, we conduct simple simulation experiments on RTE, MRPC, and CoLA. Specifically, we generate noisy training data by randomly changing a certain proportion of labels to incorrect ones. We test the robustness of different fine-tuning methods trained on the noisy data.

As shown in Tab.~\ref{table:noise}, \ours{} outperforms other fine-tuning methods on noised training data. To be exact, \ours{} surpasses vanilla fine-tuning by 1.76 average scores under 5\% noise ratio, 2.36 under 10\% noise ratio, and 1.41 under 15\% noise ratio. Furthermore, we compare the degradation of model performance towards different noise ratios. Compared with Tab.~\ref{table:main}, we calculated the degradation and display it in brackets (the last column of the Tab.~\ref{table:noise} ). It shows that \ours{} has the smallest performance drop compared to other fine-tuning methods. All the above results show that \ours{} is more robust to label noise than existing methods.

\input{float/table-time}

\subsubsection{Robustness to Data Imbalance}
\paragraph{Minority Class} Minority class refers to the class which owns insufficient instances in the training set. These kinds of classes are more challenging during fine-tuning than a normal class. To explore the performance of different tuning approaches on the minority class, we conduct experiments on synthetic RTE, MRPC, and CoLA datasets.

As Tab.~\ref{table:minority} illustrated, under different data reduction ratios, \ours{} outperforms other fine-tuning methods by a large margin. \ours{} yields an improvement of up to 6.12 average score on 30\% reduction ratio and 4.89 average scores on 40\% reduction ratio. Besides, it can be seen that the smaller the reduction ratio, the better \ours{} performs compared to other fine-tuning methods. In summary, we can conclude that \ours{} is more robust towards the minority class.\input{float/table-minority}
\paragraph{Atypical Groups} Vanilla fine-tuned models can be highly accurate on average on an i.i.d. test set yet consistently fail on atypical groups of the data~\citep{hovy2015tagging,sagawa2019distributionally} (e.g., by learning spurious correlations that hold on average but not in such groups). In contrast, \ours{} no longer aims at minimizing the original batch average loss and paying more attention to the comparison of instances. As demonstrated in Tab.~\ref{table:imbalance}, simply applying \ours{} improves the worst-group performance by 1.1 with traditional empirical risk minimization (ERM) and 1.5 with GroupDRO~\citep{sagawa2019distributionally}. What's more, the results show that \ours{} is orthogonal to prior techniques for data imbalance, integrating \ours{} with them brings further improvement.\input{float/table-imbalance}
% \subsubsection{Robustness to Spurious Bias}
% model-agnostic debiasing without prior information about bias
\section{Analysis and Discussion}

\subsection{Exploration into Effects of In-batch Instances}
As analyzed in Section \ref{sec:theory}, negative instances and positive instances in a batch function differently in the process of \ours{}. To further explore the role of negative instances and positive instances in \ours{}, we define a mask matrix $A$ by:
\begin{equation}
A_{i,j} =\left\{
\begin{array}{lcc}
1 , & \vect{y}^{(i)} = {y}^{(j)}\\
0 , & \vect{y}^{(i)} \neq {y}^{(j)}
\end{array} \right.
\end{equation}
Then we update $M \leftarrow A \odot M$ so that merely positive instances are involved for \ours{}, while $M \leftarrow (\boldsymbol{I}_{n} - A) \odot M$ so that only negative instances are observed.

\input{float/table-mask}
As is shown in Tab.~\ref{table:mask}, both negative instances and positive instances place an important role in \ours{}. When negative instances are masked for the matching matrix and only positive instances are responsible for \ours{}, the resulting score outperforms on RTE and MRPC, but drops on CoLA slightly. In the contrast, if we only preserve the influence of negative instances on the current instance, the performance surpasses the vanilla fine-tuning baseline steadily.

This result indicates that both perturbation and interpolation contribute to the final improvements. 
\ours{} simply unify negative instances and positive instances by the matching matrix, and such unification brings further improvement (refer to the last row of Tab.~\ref{table:mask}). 

\subsection{Computational Efficiency}
\label{appendix:time}
\ours improves the general performance and robustness of PLMs by introducing simple in-batch interactions.
To demonstrate the computational efficiency of \ours, we report the training time of a single epoch for different fine-tuning methods.
All the methods are based on BERT$_{\mathrm{LARGE}}$ and tested on a single NVIDIA A40 GPU.
As illustrated in Tab.\ref{table:time}, while other methods all introduce heavy extra computational cost, \ours takes almost no overhead than vanilla fine-tuning. 
Besides, \ours + R3F improves the performance of R3F with slight overhead, which also shows the efficiency of \ours.

\section{Conclusions}
% To alleviate the representation collapse problem and
To improve the general performance and robustness for fine-tuning PLMs, we propose robust \ours{} via in-batch instance perturbation. Extensive experiments on downstream tasks demonstrate the general performance of \ours{}. In addition, \ours{} is shown to be a powerful tuning approach towards broad categories of robustness evaluation. We further analyze the functioning process of \ours{} and provide probation on its components. 

\section*{Acknowledgements}
This paper is supported by the National Key Research and Development Program of China 2020AAA0106700 and NSFC project U19A2065.

\bibliographystyle{named}
\bibliography{ijcai22}
\clearpage
\appendix

\section{Statistical Information of GLUE datasets}
\label{appendix:dataset}
In this paper, we conduct experiments on datasets in GLUE benchmark~\citep{wang2018glue}. The statistic information of GLUE benchmark is shown in Table~\ref{table:glue-benchmark}.
\begin{table}[htbp]
\centering
\begin{tabular}{lccc}
\toprule
\bf Dataset & \bf \#Train & \bf \#Dev & \bf Metrics \\
\midrule
\multicolumn{3}{l}{\textit{Single-sentence Tasks}} \\
CoLA & 8.5k & 1.0k & Matthews Corr \\
SST-2 & 67k & 872 & Accuracy \\
\midrule
\multicolumn{3}{l}{\textit{Inference}} \\
RTE & 2.5k & 277 & Accuracy \\
QNLI & 105k & 5.5k & Accuracy \\
MNLI & 393k & 9.8k & Accuracy \\
\midrule
\multicolumn{3}{l}{\textit{Similarity and Paraphrase}} \\
MRPC & 3.7k & 408 & F1 \\
STS-B & 5.7k & 1.5k & Spearman Corr \\
QQP & 364k & 40k & F1 \\
\bottomrule
\end{tabular}
\caption{Statistics and metrics of eight datasets used in this paper form GLUE benchmark.
}
\label{table:glue-benchmark}
\end{table}

We also conduct experiments on advGLUE benchmark~\citep{wang2018glue} as shown in Table~\ref{table:advglue-benchmark}
\begin{table}[htbp]
\small
    \centering
\setlength{\tabcolsep}{10pt}
    \begin{tabular}{lccc}
    \toprule
      \multirow{2}{*}{\textbf{Corpus}}  & \multirow{2}{*}{\textbf{Task}} & \textbf{|Train|} & {\textbf{|Test|}}   \\   
      & & \scriptsize{(GLUE)} &  \scriptsize{(AdvGLUE)}  \\
        \midrule
        \textbf{SST-2} & sentiment & 67,349 & 1,420 \\
        \textbf{QQP}  & paraphrase & 363,846 & 422 \\
        \textbf{QNLI} & NLI/QA & 104,743 & 968 \\
        \textbf{RTE}  & NLI & 2,490 & 304 \\
        \textbf{MNLI} & NLI & 392,702 & 1,864 \\
        \midrule
        \multicolumn{3}{l}{\textbf{Sum of method test set}} & 4,978 \\
        \bottomrule
        \end{tabular}
 %   }
% \vspace{-4mm}
    \caption{Statistics  of advGLUE benchmark}
    \label{table:advglue-benchmark}
\end{table}

\section{Settings for Different Pretrained Language Models}
\label{appendix:setup}
We fine-tune different large pretrained language models with \ours{}, including
BERT$_{\mathrm{LARGE}}$\footnote{\url{https://huggingface.co/bert-large-cased/tree/main}},
and ELECTRA$_{\mathrm{LARGE}}$\footnote{\url{https://huggingface.co/google/electra-large-discriminator/tree/main}}.
The training information is listed in Table~\ref{table:settings}.

We use grid search for learning rate from $\left \{ 1\text{e-}5, 2\text{e-}5, \dots, 1\text{e-}4 \right \}$. For \ours{}, We use grid search for temperature hyperparameter from $\left \{ 1.0, 2.0, \dots, 6.0\right \}$.
We conduct all the experiments on a single A40 GPU.

\begin{table}[t]
\centering
\small
\begin{tabular}{ccccc}
\toprule
\bf Model & \bf Dataset & \bf Batch Size & \bf Epochs & Warmup Ratio \\
\midrule
BERT & all & 16 & 3 epochs & 10\% \\
\midrule
\multirow{4}{*}{ELECTRA} & CoLA & 32 & 3 epochs & 10\% \\
~ & RTE & 32 & 10 epochs & 10\% \\
~ & MRPC & 32 & 3 epochs & 10\% \\
~ & STS-B & 32 & 10 epochs & 10\% \\
\bottomrule
\end{tabular}
\caption{Hyperparameters settings for different pretrained models on variant tasks. These settings are reported in the their official repository for \emph{best practice}.}
\label{table:settings}
\end{table}

% \section{Pytorch implementation}
% Fig. \ref{fig:pytorch-losses-code} provides the simple pytorch implementation of MatchTuning.
% \begin{figure}[t]
%     \input{alg/pytorch}%
%     \caption{PyTorch implementation of \ours{}} \label{fig:pytorch-losses-code}%
% \end{figure}%

\section{Experiments on Other Pretrained Language Models}
\label{append:other}

Theoretically, our \ours method, which only adds a matching matrix on the outputs of the PLMs, can be applied on different PLMs.
Thus we conducted experiments on ELECTRA$_{\mathrm{LARGE}}$ over 10 random seeds.
Notably, we notice that the vanilla fine-tuning process of ELECTRA$_{\mathrm{LARGE}}$ is unstable, e.g., for some random seeds, Matthews correlation of CoLA task may fall to zero.
Therefore, for vanilla fine-tuning, We report both the raw averaged score of 10 seeds and averaged score with failed experiments filtered. For a fairer comparison, for other fine-tuning methods, we report the result after filtering.

As is shown in Tab.\ref{table:electra}, \ours provides an improvement of 0.57 average score on ELECTRA$_{\mathrm{LARGE}}$, which demonstrates \ours is model-agnostic and can consistently improve performance of different PLMs.
\input{float/table-electra}

\section{Comparison of Training Loss} \label{appendix:loss}
\begin{figure}[t]
	\centering
    \includegraphics[width=0.9\linewidth]{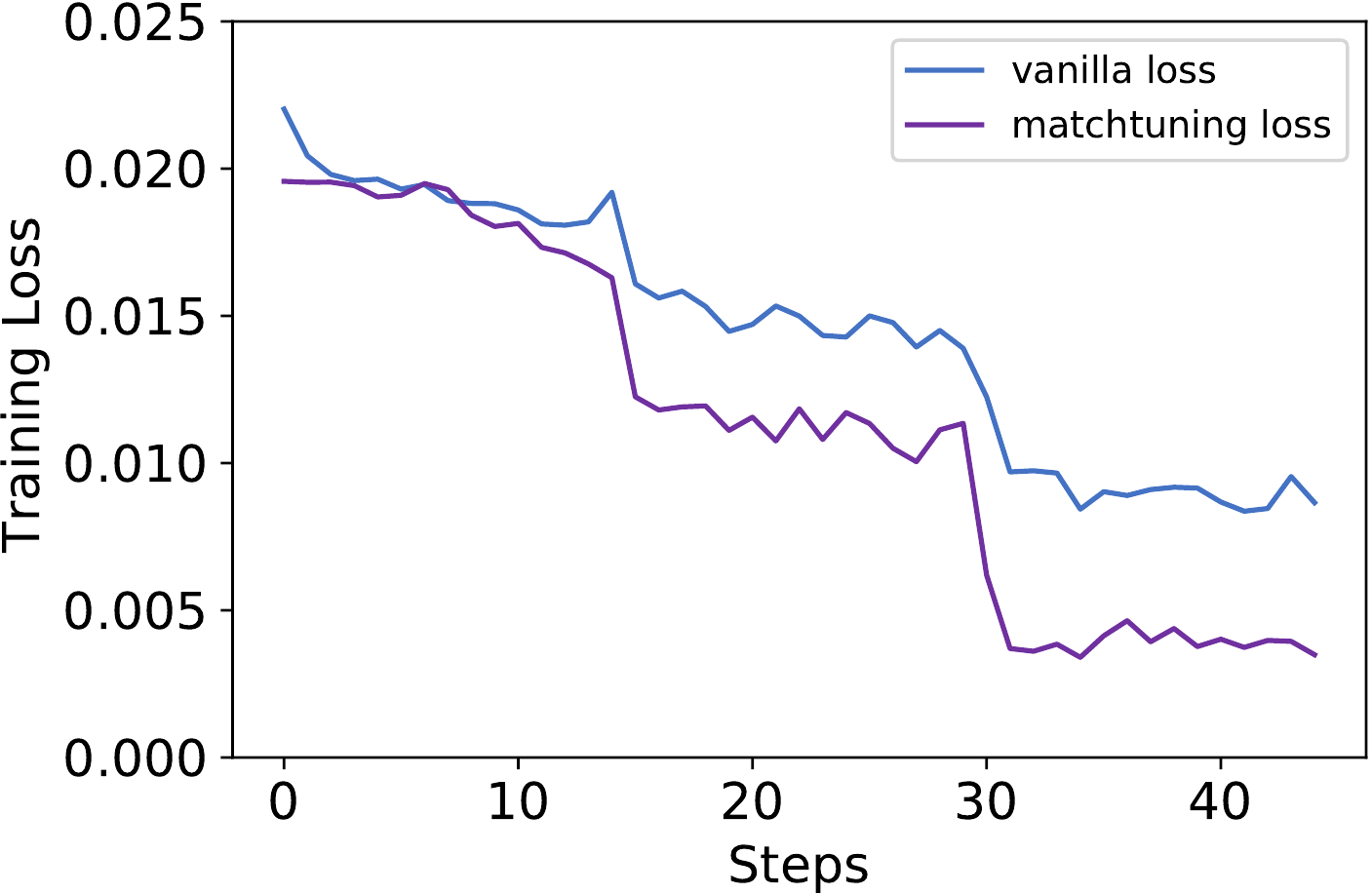}
	\caption{The comparison of training loss change on RTE between \ours{} and vanilla finetuning.}
	\label{fig:loss}
\end{figure}

The comparison of training loss change on RTE between \ours{} and vanilla finetuning. \ours{} converges faster and reaches a lower stable global minima than vanilla finetuning. For details, please refer to Fig. \ref{fig:loss}.

% \definecolor{LightGray}{gray}{0.9}

\end{document}

%% file: float/figure-process.tex
\begin{figure*}[t]
	\centering
    \includegraphics[width=0.9\textwidth]{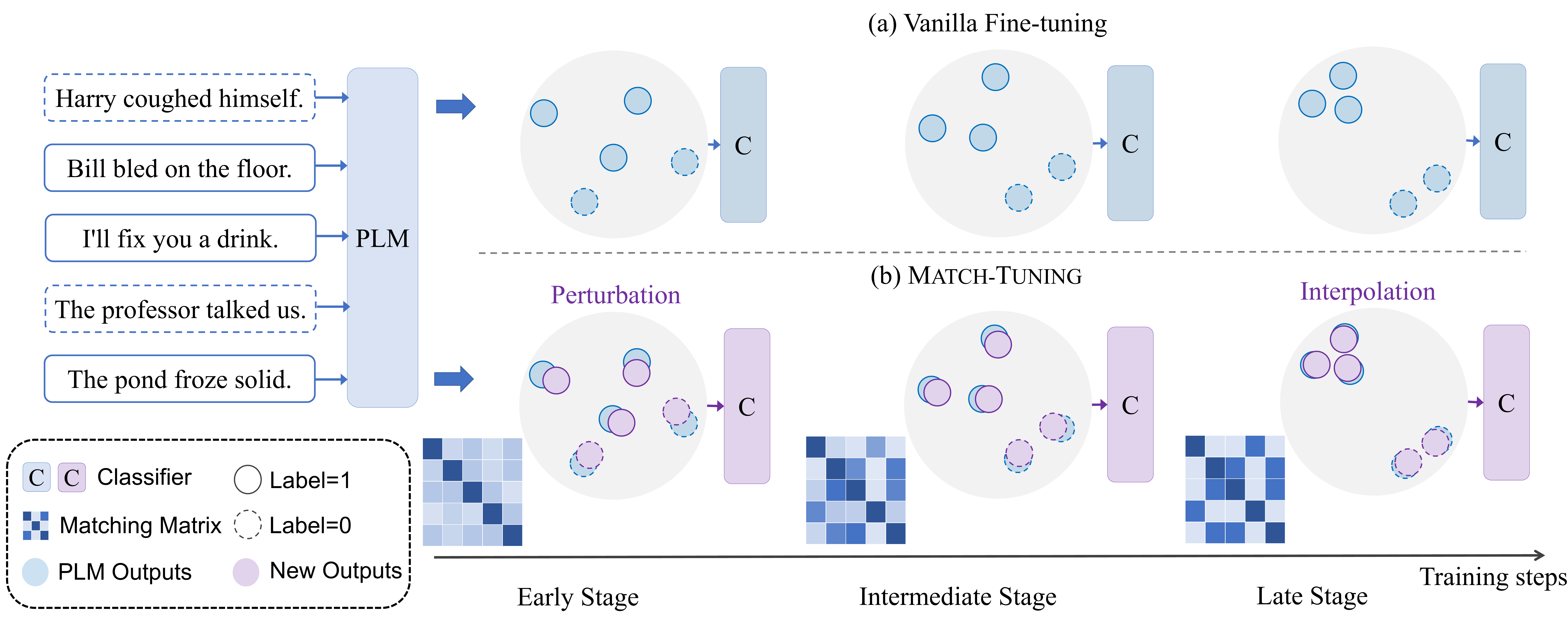}
	\caption{
% 	Comparison of \ours{} process and vanilla fine-tuning process. Sentences on the left come from the binary classification dataset, CoLA.
Illustration of \ours{}. 
Compared with the vanilla fine-tuning (a), \ours{} (b) works by adaptively determining how to utilize the in-batch instances during the training procedure. 
At first, all elements in the matching matrix except the diagonal elements are nearly random, i.e., in-batch instances work as random perturbations. 
While as the training progresses, the matrix elements corresponding to different-label instances will become smaller, and \ours{} will gradually tend to perform interpolation.
}
	\label{fig:process}
\end{figure*}

%% file: float/table-main.tex
\begin{table*}[h]
\centering
\footnotesize
\setlength{\tabcolsep}{10pt}
\begin{tabular}{@{}lcccccc@{}}
\toprule
\bf Method & \bf CoLA & \bf RTE & \bf MRPC & \bf STS-B  & \bf Avg  & $\Delta$ \\
\midrule
% Vanilla Fine-tuning[child]$^\dagger$  &      63.13 (64.31) &      70.18 (72.56) &      90.77 (91.42) &      89.61 (90.12) &   78.42   & ~~~--~~~ \\
Vanilla Fine-tuning &      63.16 (64.55) &     70.61 (74.37) &      90.70 (91.42) &     89.64 (90.99) &     78.53 &    0.00  \\
\midrule
Weight Decay~\citep{daume-iii-2007-frustratingly}  &      63.26 (64.76) &    72.10  (74.77) &    90.88 (91.62) &     89.66 (90.22) & 78.98     & +0.45      \\
Top-$K$ Tuning~\citep{pmlr-v97-houlsby19a} &     63.02 (63.88) &     70.92 (74.37) &      91.04 (92.23) &     89.64 (90.83) & 78.66    & +0.13     \\
Mixout~\citep{mixout} &    63.78 (65.55) &  72.32 (75.52) &     91.19 (92.01) &  89.89 (90.33) &   79.30   & +0.77     \\
RecAdam~\citep{RecAdam} &     63.99 (65.53) &     71.82 (73.30) &     90.84 (91.89) &     89.67 (90.42) &    79.08  & +0.55     \\
R3F~\citep{rxf} &     64.03 (66.24) &     72.42 (74.37) &    91.09 (91.32) &      ~~89.64 (90.99)$^{*}$ &    79.30  & +0.77     \\
ChildTuning$_F$~\citep{xu2021raise} &    63.70 (66.12) &      72.02 (74.17) & 91.23 (92.01) & 90.16 (90.68) &     79.28 &  +0.75     \\
ChildTuning$_D$~\citep{xu2021raise} & \bf    64.84 (66.17) &      73.23 (76.17) &      91.42 (92.20) &     90.18 (90.88) & 79.92      &  +1.39     \\
\midrule
\ours{} &  64.39 (67.25) & \bf 74.12 (76.17) & \bf      91.70 (92.39) & \bf 90.45 (90.89) & \bf  80.17   & \bf +1.64     \\
\midrule
\ours{} + R3F & \bf   65.21 (67.25) &       73.63 (76.17) & \bf     92.34 (93.22) & \bf     ~~90.45 (90.89)$^{*}$ &  \bf   80.41   & \bf +1.88     \\
\bottomrule
\end{tabular}
\caption{
Comparison between \ours{} with other fine-tuning methods.
We report the mean (max) results of $10$ random seeds. Note that since
R3F is not applicable to regression task, the results on STS-B (marked with $^{*}$ ) remain the same as vanilla and \ours{}, respectively. \ours{} achieves the best performance compared with other methods. Integrating
\ours{} with other fine-tuning methods like R3F can yield further improvements.
}
% setting: match-tuning same as main experiment; other tuning method same as child appendix C
\label{table:main}
\end{table*}

%% file: float/figure-minima.tex
\begin{figure}[t]
	\centering
    \includegraphics[width=0.38\textwidth]{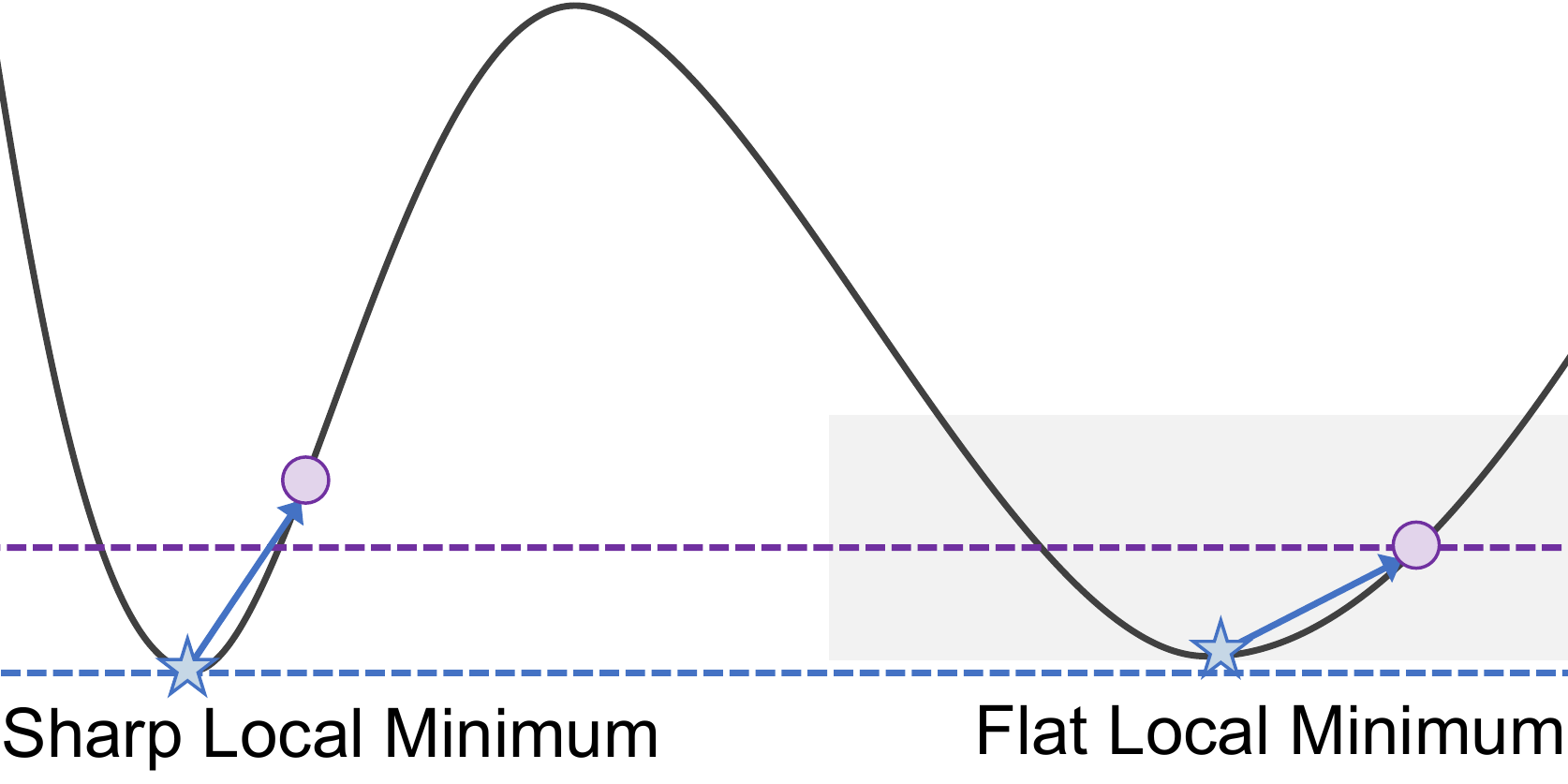}
	\caption{In this illustration of the loss function, vanilla fine-tuning prefers the sharp local minimum with lower loss rather than the flat local minimum (the blue line demonstrates the comparison). However, in \ours{}, the in-batch perturbation brings random noise and pushes the PLMs to search out the flat local minimum for better robustness and generalization.}
	\label{fig:minina}
\end{figure}

%% file: float/figure-adaptive.tex
\begin{figure}[t]
	\centering
    \includegraphics[width=0.85\linewidth]{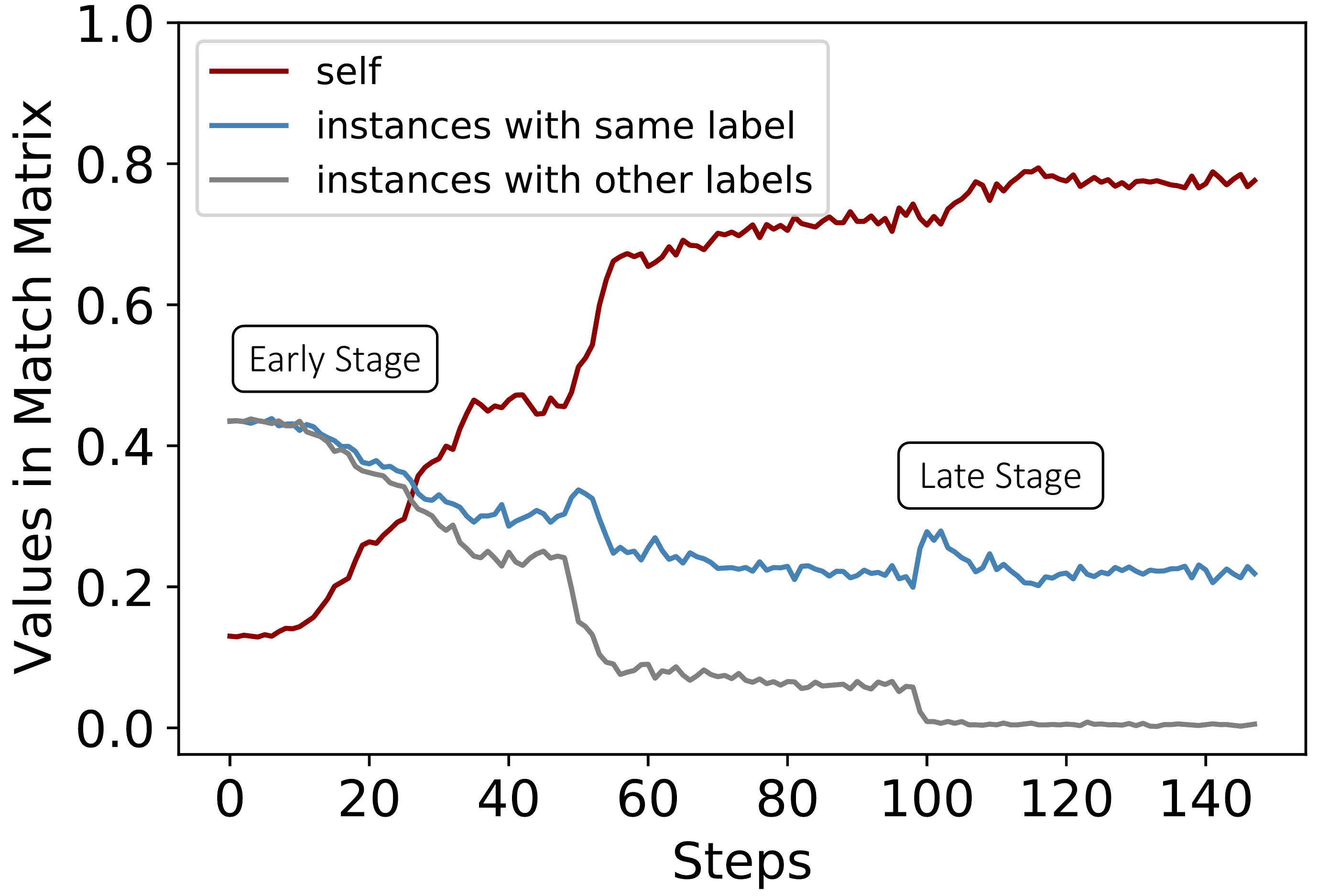}
	\caption{The change of match matrix values for the target (self), positive (same label) and negative (other labels) instances as training proceeds. The effects of \ours{} transit from in-batch perturbation to in-batch interpolation by adding adaptive noise to the instance representation. Details can be found in Sec \ref{sec:theory}.}
	\label{fig:adaptive}
\end{figure}

%% file: float/table-attack.tex
\begin{table*}[t]
\centering
\footnotesize
\setlength{\tabcolsep}{10pt}
\begin{tabular}{@{}lccccccc@{}}
\toprule
\textbf{Method} & \bf advSST-2 & \bf advMNLI & \bf advRTE & \bf advQNLI & \bf advQQP & \bf Avg &\bf$\Delta$\\
\midrule
Vanilla Fine-tuning & 47.57  &  34.99/30.00  & 41.73  &  46.40  & 38.45/27.59 & 40.24 & 0.00\\
\midrule
R3F~\citep{rxf} &  38.51  & \textbf{35.81}/30.26  &  50.12  &  47.52  & 40.59/35.23 & 41.42 &+1.18\\
$\text{ChildTuning}_F$~\citep{xu2021raise} & 34.46 &33.88/26.53  & 41.98  &47.53 & 40.38/35.82 &38.46 &-1.78\\
$\text{ChildTuning}_D$~\citep{xu2021raise} &39.19 &34.06/27.84 &46.17  &\bf49.55 & 40.66/\bf39.80 & 41.22 &+0.98\\
\ours &\bf51.35  &  35.54/\bf31.07  & \bf52.52  &  47.52  & \textbf{41.45} /32.62 & \bf44.35 &\bf+4.11\\
\bottomrule
\end{tabular}
\caption{Robustness evaluation on AdvGLUE validation set. We report the mean results of $3$ random seeds. 
% Following ~\citet{wang2021adversarial}, for MNLI, we report the test accuracy on the matched and mismatched validation sets; for QQP, we report accuracy and F1; and for other tasks, we report the accuracy. The best results are bold. 
\ours achieves considerable improvement on most datasets, especially on the SST-2 and RTE datasets. 
% And we get the largest average score and average boost compared to other methods.
}

\label{table:attack}
\end{table*}

%% file: float/table-noise.tex
\begin{table}[t]
\centering
\footnotesize
\setlength{\tabcolsep}{6pt}
\begin{tabular}{@{}lcccc@{}}
\toprule
\bf Method & \bf CoLA & \bf MRPC & \bf RTE & \bf Avg\\
\midrule
Noise Ratio & \multicolumn{4}{c}{5 \%}\\
\cmidrule(lr){1-1} \cmidrule(lr){2-5}
Vanilla Fine-tuning &61.14 & 90.38 &69.68 &73.73 ($\downarrow$ 4.80)\\
R3F &\bf62.42  &90.82  &67.99  &73.74 ($\downarrow$  5.56)\\
$\text{ChildTuning}_F$ &61.13 &90.46 &71.59 &74.39 ($\downarrow$  4.89)\\
$\text{ChildTuning}_D$ &61.46  &90.42 &71.72 &74.53 ($\downarrow$ 5.39)\\
\ours{} &62.33 &\bf91.19 &\bf72.96 & \textbf{75.49} ($\downarrow$ \textbf{4.68})\\
\midrule
Noise Ratio & \multicolumn{4}{c}{10 \%}\\
\cmidrule(lr){1-1} \cmidrule(lr){2-5}
Vanilla Fine-tuning &59.21&88.90&68.34&72.15 ( $\downarrow$ 6.38)\\
R3F &\bf61.76 &90.36 &66.75 &72.96 ($\downarrow$      6.34)\\
$\text{ChildTuning}_F$ &60.97  &89.83 &70.02 &73.61 ($\downarrow$ 5.67)\\
$\text{ChildTuning}_D$ &61.35  &89.92 &69.06 &73.44 ($\downarrow$ 6.84)\\
\ours{} &61.41 &\bf90.55 &\bf71.56 & \textbf{74.51} ($\downarrow$ \textbf{5.66})\\
\midrule
Noise Ratio & \multicolumn{4}{c}{15 \%}\\
\cmidrule(lr){1-1} \cmidrule(lr){2-5}
Vanilla Fine-tuning &59.01 &87.84 &68.12 &71.66 ($\downarrow$ \textbf{6.87})\\
R3F &\bf60.16 &88.51 &65.14 &71.27 ($\downarrow$  8.03)\\
$\text{ChildTuning}_F$ &59.66 &88.08 &69.10 &72.28 ($\downarrow$  7.00)\\
$\text{ChildTuning}_D$ &59.88 &89.01 &69.78 &72.89 ($\downarrow$  7.03)\\
\ours{} &59.65  &\bf89.51  &\bf70.04 &
\textbf{73.07} ($\downarrow$ 7.10)\\
\bottomrule
\end{tabular}
\caption{Comparison of different tuning approaches on robustness towards label noise. The noise ratio refers to the proportion of training instances whose labels are transferred to incorrect labels. \ours{} can maintain more robust representations compared with other fine-tuning methods.}
%\caption{Comparison of different tuning approaches on robustness towards label noise. The noise ratio refers to the proportion of training instances whose labels are transferred to incorrect labels. The parameters of \ours{} are same as main experiment and that of other tuning method are same as Appendix \ref{appendix:setup}.}
\label{table:noise}
\end{table}

%% file: float/table-time.tex
\begin{table*}[htbp]
\centering
\footnotesize
\setlength{\tabcolsep}{5.8pt}
\begin{tabular}{@{}lcccccccccc@{}}
\toprule
{\bf Method} & \bf CoLA &\bf RTE &\bf MRPC &\bf STS-B &\bf advSST-2 &\bf advMNLI &\bf advRTE &\bf advQNLI &\bf advQQP &\bf Avg\\
\midrule
Vanilla Fine-tuning & 80  &40  &37  &55  &645  &4090 &40  &1190  &3540 &1070\\
\midrule
R3F & 135   & 75   & 65  &- &860  &7020  &75  &2010  &5490 &1755$^{*}$\\
ChildTuning$_F$ &150  &60  &60  &100   &1200  & 6570  &60  &1950  &6600 &1861\\
ChildTuning$_D^\dagger$ &100   &45  &40  &65  &700  &4310  &45  &1270  &3720 &1144\\
\ours &\bf80 &\bf40 &\bf37 &\bf55 &\bf650  &\bf4280 &\bf40 &\bf1200 &\bf3540 &\bf1102\\
\midrule
\ours + R3F &140 &80 &65  &-  &880  &7100  &80  &2030  &5500 &1772$^{*}$\\
\bottomrule
\end{tabular}
\caption{Training time (second) for different fine-tuning methods on different datasets. We report the time of a single epoch.
Results with $^\dagger$ should be noted because ChildTuning$_D$ requires an extra epoch to calculate Fisher information compared with other fine-tuning methods. Besides, since
R3F is not applicable to the regression task, the result marked with $^{*}$ is calculated by using the average of the column STS-B. \ours requires almost no extra computation overhead compared with vanilla fine-tuning.
}
\label{table:time}
\end{table*}

%% file: float/table-minority.tex
\begin{table}[t]
\centering
\footnotesize
\setlength{\tabcolsep}{6pt}
\begin{tabular}{@{}lcccc@{}}
\toprule
{\bf Method} & \bf CoLA &\bf MRPC &\bf  RTE &\bf Avg\\
\midrule
Reduction Ratio & \multicolumn{4}{c}{30 \%}\\
\cmidrule(lr){1-1} \cmidrule(lr){2-5}
Vanilla Fine-tuning &81.11 &78.53  &27.17 & 62.27 (\quad--\quad)\\
R3F &80.50 &80.17 &20.61 &60.42 ($\downarrow$ 1.85)\\
ChildTuning$_F$ &83.07  &80.36 &29.31 &64.25 ($\uparrow$ 1.98)\\
ChildTuning$_D$ &\bf83.18 &78.75 &29.62 &63.85 ($\uparrow$ 1.58)\\
\ours &81.68 &\bf 83.26 &\bf 40.23 & \textbf{68.39} ($\uparrow$ \textbf{6.12})\\
\midrule
Reduction Ratio & \multicolumn{4}{c}{40 \%}\\
\cmidrule(lr){1-1} \cmidrule(lr){2-5}
Vanilla Fine-tuning &83.13 & 83.94 &38.14 &68.40 (\quad--\quad)\\
R3F &83.01 &85.14 &22.43 &63.53 ($\downarrow$ 4.87)\\
ChildTuning$_F$ &85.10 &82.80 &37.79 &68.56 ($\uparrow$ 0.16)\\
ChildTuning$_D$ &\bf85.76 &84.41 &38.78 &69.65 ($\uparrow$ 1.15)\\
\ours &83.30 &\bf88.03 &\bf48.55 &\textbf{73.29} ($\uparrow$ \textbf{4.89})\\
\midrule
Reduction Ratio & \multicolumn{4}{c}{50 \%}\\
\cmidrule(lr){1-1} \cmidrule(lr){2-5}
Vanilla Fine-tuning &86.01 &87.56 &44.81 &72.79 (\quad--\quad)\\
R3F &86.27 &89.21 &30.53 &68.67 ($\downarrow$ 4.12)\\
ChildTuning$_F$ &\bf88.09  &87.06 &48.58 &74.58 ($\uparrow$ 1.79)\\
ChildTuning$_D$ &87.89 &87.63 &49.41 &74.98 ($\uparrow$ 2.19)\\
\ours &86.22 &\bf89.32 &\bf53.66 &\textbf{76.40} ($\uparrow$ \textbf{3.61})\\
\bottomrule
\end{tabular}
%\caption{To conduct experiments on minority class robustness, we reduce the number the instances labelled $1$ in the training set to 30\%/40\%/50\% of the original number, and test the accuracy of instances labelled $1$ (as the minority class) in the validation set. The parameters of \ours{} are same as main experiment and that of other tuning method are same as Appendix \ref{appendix:setup}.}
\caption{To conduct experiments on minority class robustness, we reduce the number the instances labelled $1$ in the training set to 30\%/40\%/50\% of the original number, and test the accuracy of instances labeled $1$ (as the minority class) in the validation set. \ours outperforms other methods by a large margin at any reduction ratio.}

\label{table:minority}
\end{table}

%% file: float/table-imbalance.tex
\begin{table}[t]
\centering
\footnotesize
\setlength{\tabcolsep}{4pt}
\begin{tabular}{@{}lcc@{}}
\toprule
\multirow{2}{*}{\bf Method} & \multicolumn{2}{c}{\bf MNLI} \\
\cmidrule(lr){2-3}   
~ & Avg Acc & Worst-group Acc  \\
\midrule
ERM &  82.8 (~~~--~~~) & 65.1 (~~~--~~~) \\
GroupDRO &  81.2 (~~~--~~~) &  78.3 (~~~--~~~) \\
ERM + \ours{} & 82.9 ($\uparrow$ 0.1) &  66.1 ($\uparrow$ 1.0) \\
GroupDRO + \ours{} & 81.2 ($\uparrow$ 0.0) & 79.6 ($\uparrow$ 1.3)\\
\bottomrule
% ERM &  82.9 & 76.0 \\
% GroupDRO &  84.0 &  80.4 \\
% ERM + \ours & 84.9 &  78.1\\
% GroupDRO + \ours & 84.2 & 81.3 \\
% \bottomrule
\end{tabular}
\caption{We conduct group robustness evaluation on MNLI in GLUE. Following previous work~\citep{sagawa2019distributionally}, we divide the MNLI dataset into six groups, one for each pair of labels in \{entailed, neutral, contradictory\} and spurious attributes in \{no negation, negation\}. 
% imbalance: bert-base-uncased above, bert-large-uncased below. 
% setting: keep the same as the dro thesis; match-tuning same as main experiment. 
}
\label{table:imbalance}
\end{table}

%% file: float/table-mask.tex
\begin{table}[t]
\centering
\footnotesize
\setlength{\tabcolsep}{5pt}
\begin{tabular}{@{}lccc@{}}
\toprule
\bf Method & \bf CoLA & \bf RTE & \bf MRPC  \\
\midrule
F.T.$_{vanilla}$ & 63.16 ($\quad$--$\quad$) &  70.61  ($\quad$--$\quad$) & 90.70  ($\quad$--$\quad$) \\
M.T.$_{Positive}$ &  62.88 ($\downarrow$ 0.28) &  72.71 ($\uparrow$ 2.10) & 90.77 ($\uparrow$ 0.07) \\
M.T.$_{Negative}$ &  63.32 ($\uparrow$ 0.16) &  72.82 ($\uparrow$ 2.21) & 91.18  ($\uparrow$ 0.48)\\
M.T.$_{Full}$ &  64.39  ($\uparrow$ 1.23) &  74.12 ($\uparrow$ 3.51)  &  91.70 ($\uparrow$ 1.00) \\
\bottomrule
\end{tabular}
\caption{Use masking strategies to explore the role of positive and negative instances. F.T. and M.T. are the abbreviations for Fine-tuning and \ours{}, respectively. It indicates that both positive and negative instances contribute to the final enhancements.}
\label{table:mask}
\end{table}

%% file: float/table-electra.tex
\begin{table}[hbpt]
\centering
\footnotesize
\setlength{\tabcolsep}{2pt}
\begin{tabular}{@{}lccccc@{}}
\toprule
\bf Method & \bf CoLA & \bf RTE & \bf MRPC & \bf STS-B & \bf Avg  \\
\midrule
ELECTRA+Vanilla$_{raw}$  &49.19  &88.18  &92.84  &82.27  &78.12 \\
ELECTRA+Vanilla$_{filtered}$ &70.52 &88.18 &92.84 &91.62 & 85.76 \\
ELECTRA+ChildTuning$_F$ &70.77 &88.62 &93.01 &91.76 &86.04\\ 
ELECTRA+ChildTuning$_D$ &70.98 &88.94 &\bf93.21 &91.92 &86.26\\ 
ELECTRA+\ours &\bf71.19 &\bf88.97 &93.17 &\bf 91.97  &\bf 86.33 \\
\bottomrule
\end{tabular}
\caption{Comparison between \ours{} and vanilla fine-tuning on ELECTRA$_{\mathrm{LARGE}}$. \ours{} outperforms other fine-tuning methods}
% setting: lr: [1e-5, 2e-5, ..., 1e-4], temperature: [1, 2, ..., 8]. All other parameters follow the recommended parameters in the original papers of each model. You can refer to: Table 7 in the appendix of child-tuning
\label{table:electra}
\end{table}